\documentclass[runningheads]{llncs}
\usepackage{graphicx}
\usepackage{tikz}
\usepackage{comment}
\usepackage{amsmath,amssymb} % define this before the line numbering.
\usepackage{color}
\usepackage{xcolor}
\usepackage{cite}

\definecolor{brightlavender}{rgb}{0.79, 0.6, 1.0}
\definecolor{brickred}{rgb}{0.89, 0.25, 0.23}
\definecolor{applegreen}{rgb}{0.64, 0.82, 0.63}

\definecolor{forestgreen}{rgb}{0.0, 0.5, 0.0}

\usepackage[accsupp]{axessibility}  % Improves PDF readability for those with disabilities.

\begin{document}
% \renewcommand\thelinenumber{\color[rgb]{0.2,0.5,0.8}\normalfont\sffamily\scriptsize\arabic{linenumber}\color[rgb]{0,0,0}}
% \renewcommand\makeLineNumber {\hss\thelinenumber\ \hspace{6mm} \rlap{\hskip\textwidth\ \hspace{6.5mm}\thelinenumber}}
% \linenumbers
\pagestyle{headings}
\mainmatter
\def\ECCVSubNumber{8001}  % Insert your submission number here

\title{Bridging the visual semantic gap in VLN via semantically richer instructions} % Replace with your title

% INITIAL SUBMISSION 
\begin{comment}
\titlerunning{ECCV-22 submission ID \ECCVSubNumber} 
\authorrunning{ECCV-22 submission ID \ECCVSubNumber} 
\author{Anonymous ECCV submission}
\institute{Paper ID \ECCVSubNumber}
\end{comment}
%******************

% CAMERA READY SUBMISSION
%\begin{comment}
\titlerunning{Semantically richer instructions for VLN}
% If the paper title is too long for the running head, you can set
% an abbreviated paper title here
%
\author{Joaquín Ossandón\orcidID{0000-0003-3122-0631} \and
Benjamín Earle\orcidID{0000-0003-4391-6767} \and
Álvaro Soto\orcidID{0000-0001-9378-397X}}
\authorrunning{J. Ossandón et al.}
% First names are abbreviated in the running head.
% If there are more than two authors, 'et al.' is used.
%
\institute{
Pontificia Universidad Católica de Chile, Santiago, Chile\\
\email{\{jiossandon,biearle\}@uc.cl}\\
\email{asoto@ing.puc.cl}}
%\end{comment}
%******************
\maketitle

\begin{abstract}
The Visual-and-Language Navigation (VLN) task requires understanding a textual instruction to navigate a natural indoor environment using only visual information. While this is a trivial task for most humans, it is still an open problem for AI models. In this work, we hypothesize that poor use of the visual information available is at the core of the low performance of current models. To support this hypothesis, we provide experimental evidence showing that state-of-the-art models are not severely affected when they receive just limited or even no visual data, indicating a strong overfitting to the textual instructions. To encourage a more suitable use of the visual information, we propose a new data augmentation method that fosters the inclusion of more explicit visual information in the generation of textual navigational instructions. Our main intuition is that current VLN datasets include textual instructions that are intended to inform an expert navigator, such as a human, but not a beginner visual navigational agent, such as a randomly initialized DL model. Specifically, to bridge the visual semantic gap of current VLN datasets, we take advantage of metadata available for the Matterport3D dataset that, among others, includes information about object labels that are present in the scenes. Training a state-of-the-art model with the new set of instructions increase its performance by 8\% in terms of success rate on unseen environments, demonstrating the advantages of the proposed data augmentation method.
\end{abstract}

\section{Introduction} \label{intro}

The ability of a robot to receive an instruction in natural language and navigate in unknown environments has been an attractive research topic in recent years \cite{speaker-follower, ActiveVisGathering, environmental-dropout, ManterolaValenzuelaRaimundo2021Evnb, reasoning-tasks, text-pairs}. In particular, the Visual-and-Language Navigation (VLN) task \cite{R2RPaper} proposes that an agent can follow a textual instruction such as ``\textit{Go up the stairs, turn right, and stop right at the left of the table}", and use it to navigate a natural indoor environment from a starting to a goal position using only visual information. In spite of current advances in AI, this task, that results trivial for most humans, it is still out of reach for autonomous robots. As an example, under current benchmarks \cite{survey}, state-of-the-art AI models based on Deep Learning (DL) do not reach the intended goal position more than 65\% of the time \cite{vln-bert-rem}. 

There are several reasons that can help to explain the low performance of current models to face the VLN task \cite{survey}. Among them, we believe that lack of a proper visual understanding of the environment is a key factor. In effect, humans actively use relevant views of the environment to identify visual semantic information such as navigational cues, objects, scenes, or other situations, however, current AI models focus their operation on identifying relevant correlations between the textual instructions and visual data present in the training set \cite{GaryMarcus:2018}. As a consequence, current VLN models exhibit limited generalization capabilities, leading to a large drop in performance when they are tested in unseen environments \cite{speaker-follower, environmental-dropout, survey}. 

In effect, today there is abundant experimental evidence indicating that current DL based models operate as associative memory engines triggered by superficial data correlations \cite{DL-SurfacePatterns:Bengio:2017,Memory:YBengio:ICML:2017,Belkin:Neurips:2018}, fostering the detection of direct stimulus-response associations. Indeed, given enough parameters, DL models are able to
memorize arbitrary noisy data \cite{Memorization:ICLR:2017}. In the case of VLN, this problem leads to a poor use of the visual information. As a consequence, instead of unveiling the richness of the visual world, DL models limit their operation to memorize low level correlations between textual and visual data. Even worse, in several cases, models ignore completely the visual information, learning a direct mapping between the textual instructions and robot action. 

To support the previous observation, as a first contribution of this work, we provide experimental evidence indicating that current VLN models do not make a suitable use of the visual information available about the environment. Specifically, we demonstrate that when we provide to the model just limited or even no visual data, the model exhibits just a slight drop in performance, showing that their operation is heavily biased to the use of textual instructions.  

The previous observation motivates our main research question: how can we contribute to improve the use visual information in VLN models?. While the answer to this question is manifold, in this work we focus our contribution to the generation of more suitable training data. Specifically, we believe that a relevant problem of current VLN datasets is that, during their generation, the humans providing the textual instructions assume that they are intended for an expert navigator, as an example, another human. We believe that this scheme leads to the generation of high level textual instructions, where it is hardly complex to extract meaningful visual cues to inform a beginner visual navigational agent, such as a randomly initialized DL model. As a consequence, we believe that the data generation for a beginner should include a more detailed description of the visual world around the agent. 

To bridge the visual semantic gap of current VLN datasets, we present a new data augmentation method that fosters the inclusion of more explicit visual information in the generation of textual navigational instructions. To do this, we resort to object labels present in the metadata available for the Matterport3D dataset \footnote{https://github.com/niessner/Matterport/tree/master/metadata} that we refer here as Matterport3DMeta. Using this data, we propose new semantically richer natural language instructions for the Room-to-Room (R2R) dataset \cite{R2RPaper} that are generated with an improved version of the Speaker-Follower model presented in \cite{speaker-follower}. Specifically, we use scene objects and crafted instructions created with a set of rules that we encode to feed a set of auxiliary visual tasks. As a main finding, the resulting navigational instructions provide a significant boost in the performance of current VLN models when they are tested in previously unseeing environments.

As a further contribution, after publication, we will make available the semantically enriched dataset generated in this work as well as a set of software tools to generate further data. These tools incorporate modules to access scene nodes in the Matterport3D dataset \cite{Matterport3D} that include information about relevant objects, their position, size, distance, heading, and elevation. We believe that this is a powerful starting point to use scene metadata to create semantically richer visual navigational instructions.

This work is organized as follows. Section \ref{Task} describes the VLN task and current benchmarks. Section \ref{stateoftheart} reviews relevant previous works. Section \ref{problem} presents an experimental setup to highlight the limitations of current VLN models to use visual information. Afterwards, Section \ref{section:object-based} describes the construction of our visual semantically richer instructions for the VLN task. Finally, Section \ref{conclusions} presents our conclusions and future research avenues.

\section{Visual-and-Language Navigation task} \label{Task}

During the last decade several studies have been related to the VLN task, however, the visual aspect was discarded due to lack of real images in the proposed problems \cite{R2RPaper}. In 2017, the Matterport3D dataset \cite{Matterport3D} was introduced, containing RGB-D building scale scenes of 90 different home environments. Later that year, a new navigation problem was proposed: Room-to-Room (R2R) \cite{R2RPaper}, the first dataset for the Visual-and-Language Navigation task (VLN) on real 3D environments, introducing a Matterport3D based simulator, which simulates its environments with the possibility of navigate trough them. In R2R, 90 different environments from Matterport3D have been divided into training and validation (seen and unseen) splits. There are a total of 7,189 distinct paths (starting point, target point), with 3 distinct human instructions for each, a total of 21,567 navigation instructions with an average of 29 words \cite{R2RPaper}.

\begin{figure*}[t]
    \centering
    \includegraphics[width=\linewidth]{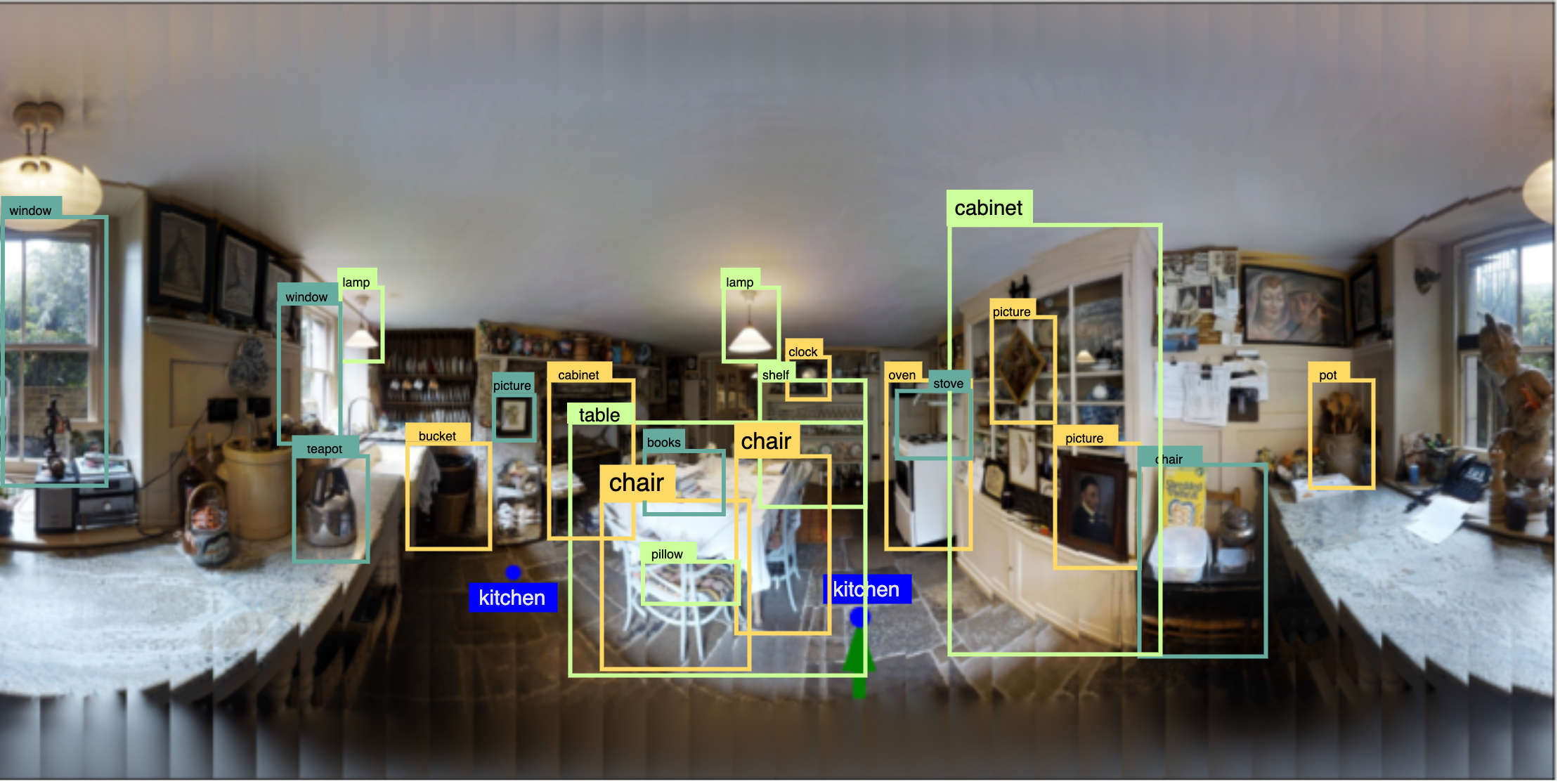}
    \caption{Objects and viewpoints visualization sampled with \texttt{360-visualization} scripts.}
    \label{fig:360-visualization}
\end{figure*}

We construct over Matterport3DMeta a set of tools named  \texttt{360-visualization}\footnote{\url{https://github.com/cacosandon/360-visualization}} for getting objects and navigable nodes with their intrinsic data for each view, as shown in Figure \ref{fig:360-visualization}.

\subsection{The Visual-and-language navigation task} \label{r2r}
The task of VLN for an agent is to follow natural language instructions from an initial to a target position through navigation in a real environment, simulated by Matterport3D Simulator \cite{R2RPaper}. At the beginning of each episode an
instruction $\overline{x} = \langle x_1, x_2, .., x_L \rangle$ is given, where $L$ is the instruction length and $x_i$ a word token. The agent observes a RGB image $v_0$ depending on an initial 3D position, heading $\psi_0$ and elevation $\theta_0$, resulting in a world state $s_0 = \langle v_0, \psi_0, \theta_0 \rangle$. The agent must execute a sequence of actions $\langle a_0, a_1, .., a_T \rangle$ where each action $a_t$ leads to a new state $s_{t+1} = \langle v_{t+1}, \psi_{t+1}, \theta_{t+1} \rangle $ and generates a new visual panoramic view $v_{t+1}$. It is important to note that actions are given by the simulator, which are limited according to the node where the agent is located. The episode ends when the agent selects the \texttt{<STOP>} action, and the task is successful if the agent arrives at a location near the target position, recognizing it as the goal.

\section{Related Work} \label{stateoftheart}

VLN task has been the main motivation for many researchers on Computer Vision. Interesting surveys and reviews \cite{survey, looking} talk about several techniques developed over the baseline architecture proposed by R2R. 

They group them on categories such as the inclusion of auxiliary tasks \cite{Self-Monitoring, reasoning-tasks, RoadToKnowWhere, ManterolaValenzuelaRaimundo2021Evnb, LXMERT}, the improvement of navigation and exploration \cite{self-backtracking, regretful, Self-Monitoring, SSM, ActiveVisGathering} and curriculum learning with data augmentation \cite{speaker-follower, text-pairs, environmental-dropout}.

Data augmentation has become an essential part of training in various tasks, not only increasing quantity of training data, but also providing more informative data to reduce overfitting, and improve generalization and performance \cite{ShortenK19, data-augmentation-images, data-augmentation-nlp-survey}. On navigation, different approaches have proposed augmenting training instructions \cite{speaker-follower, environmental-dropout} but it has been shown that they do not follow human syntax or include relevant information \cite{zhao2021evaluation}. 

That's why we focus on this topic, basing our study on the Speaker-Follower \cite{speaker-follower}, which consists of two modules: one that follows instructions (follower) and other that performs data augmentation to feed the training of the follower (speaker), which we improve for generate new semantically richer instructions.

State-of-art leaderboard is summarized in Table \ref{table:comparison}. VLN-BERT+REM \cite{vln-bert-rem} has the highest success rate (SR), followed by SSM \cite{SSM} and Active Gathering \cite{ActiveVisGathering}. These models have high overhead costs due to the time and resources required by their complex architectures. We demonstrate that focusing on data augmentation greatly benefits navigation performance without making models even more complex.

\setlength{\tabcolsep}{4pt}
\begin{table}
\begin{center}
\caption{Comparison of the different models solving the Room-to-Room task, in unseen test set using Single Run.}
\label{table:comparison}
\begin{tabular}{lllll}
\hline\noalign{\smallskip}
Model & PL $\downarrow$ & NE $\downarrow$ & SPL $\uparrow$ & SR $\uparrow$ \\
\noalign{\smallskip}
\hline
\noalign{\smallskip}
Speaker-Follower \cite{speaker-follower} & 14.82 & 6.62 & 0.28 & 0.35 \\
Tactical Rewind \cite{self-backtracking} & 22.08 & 5.14 & 0.41 & 0.54 \\
Self-Monitoring \cite{Self-Monitoring} & 17.11 & 5.99 & 0.32 & 0.43 \\
Environmental Dropout \cite{environmental-dropout} & 11.70 & - & 0.47 & 0.51 \\
Regretful-Agent \cite{regretful} & 13.69 & 5.69 & 0.40 & 0.48 \\
ORIST \cite{RoadToKnowWhere} & 10.90 & 4.72 & 0.51 & 0.57 \\
VLN-BERT + REM \cite{vln-bert-rem} & 13.11 & 3.87 & 0.59 & \textbf{0.65} \\
SSM \cite{SSM} & 20.7 & 4.32 & 0.45 & 0.62 \\
Active Gathering \cite{ActiveVisGathering} & 20.6 & 4.36 & 0.4 & 0.58 \\
\hline
\end{tabular}
\end{center}
\end{table}
\setlength{\tabcolsep}{1.4pt}

\section{Models problem} \label{problem}
\begin{figure*}[h]
    \centering
    \includegraphics[width=0.6\linewidth]{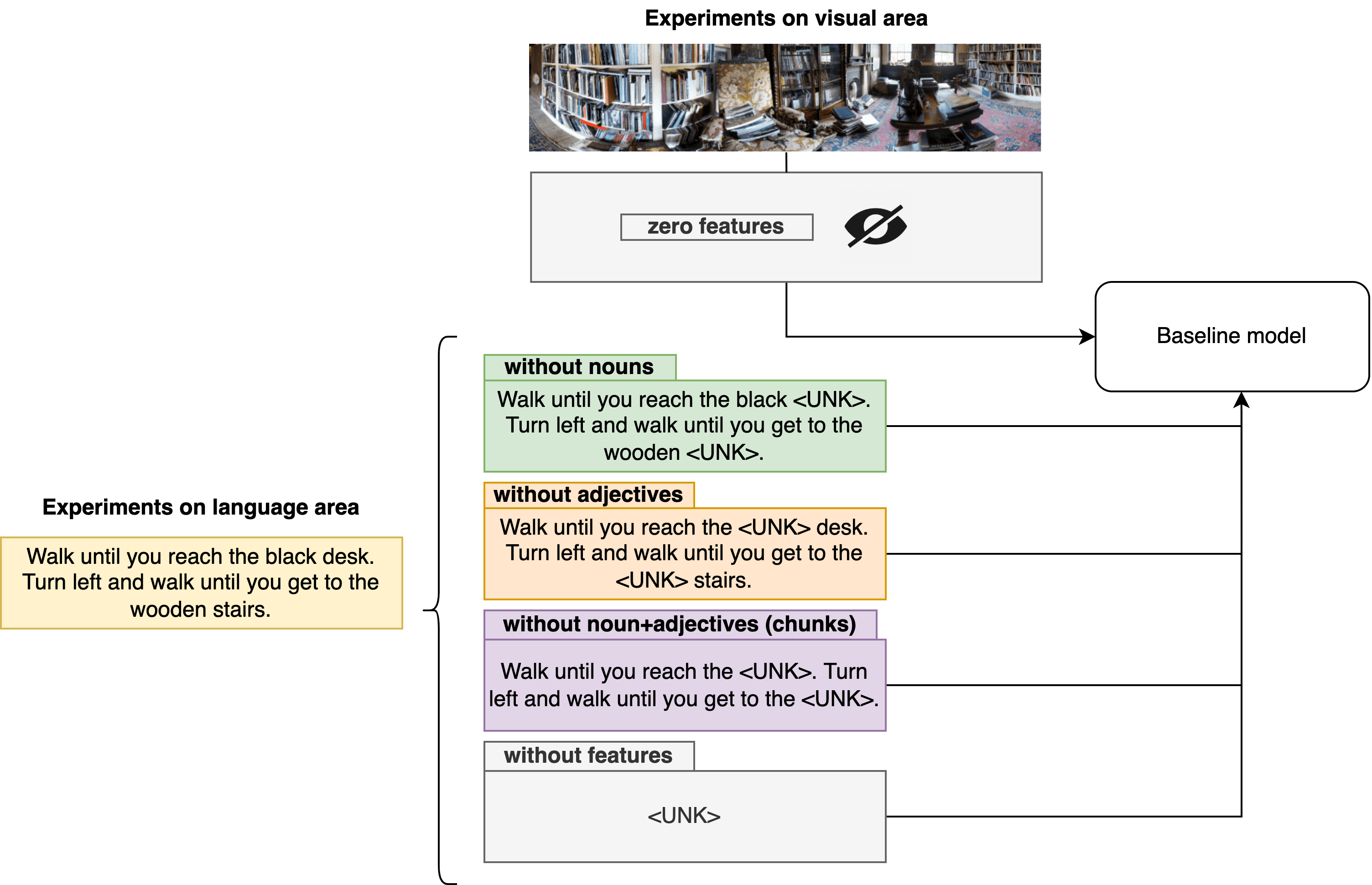}
    \caption{Experiments diagram. Visual features are replaced with zeros. In the linguistic area, four experiments are performed: without nouns, without adjectives, without nouns+adjectives and without features.}
    \label{fig:diagram}
\end{figure*}

Aiming to demonstrate models deficits, we experiment in both visual and linguistic areas on the state-of-the-art models, based on \cite{looking} analysis. A summary diagram is shown in Figure \ref{fig:diagram}.

\subsection{Visual area}
In order to evaluate the effectiveness of visual components within VLN architectures, it is interesting to know the importance of visual scene information when deciding the next action on navigation through the environment.

We run our experiments over Self-Monitoring \cite{Self-Monitoring} and Regretful-Agent \cite{regretful}, because of their public codebase\footnote{\url{https://github.com/chihyaoma/selfmonitoring-agent/}} \footnote{\url{https://github.com/chihyaoma/Regretful-Agent}}. Each of these architectures were trained in two different conditions. The first condition is the base model, where the visual features are obtained from a pre-trained ResNet-152. The second condition is the replacement of visual features with zeros, i.e., the agent is completely blind.

Because both models are built on top of the Speaker-Follower architecture, they also offer an optional pre-training phase that includes training with synthetic data. This synthetic data contains 178,000 sampled routes with associated instructions generated with the Speaker module \cite{speaker-follower}, the same instructions we improve later in this work. Six experiments evaluated in known (seen) and unknown (unseen) environments were performed, which are shown in Tables \ref{table:blind-study-seen} and \ref{table:blind-study-unseen}. 

\subsection{Language area}

We also experiment changing the text of the instruction, in order to check which components are relevant for the agent to better decide. 

We use spaCy \cite{spacy2}, an NLP model used in the industry to obtain text features. Each word of each instruction was classified according to the context, as adjective, noun or other. The Regretful-Agent model was trained by extracting from each instruction: all adjectives, all nouns, all nouns+adjectives and extracting the whole text (i.e. without linguistic features), training the model for a total of 100 hours with the 8 experiments, which results are shown in Table \ref{table:text-study-unseen}.

\subsubsection{Comparison metrics} \label{metrics}
To compare the performance of presented configurations, we use path length (PL), navigation error (NE) and success rate (SR), as proposed in R2R \cite{R2RPaper}. We also use a new metric called success rate weighted by Path Length (SPL) \cite{text-pairs, environmental-dropout, reasoning-tasks, Self-Monitoring}, that measures the success rate normalized by path length.

\setlength{\tabcolsep}{4pt}
\begin{table}
\begin{center}
\caption{Visual ablation study on Self-Monitoring \cite{Self-Monitoring} and Regretful-Agent \cite{regretful}, seen environment with Single Run (not Beam Search).}
\label{table:blind-study-seen}
\begin{tabular}{lllll}
\hline\noalign{\smallskip}
Model & PL $\downarrow$ & NE $\downarrow$ & SPL $\uparrow$ & SR $\uparrow$ \\
\noalign{\smallskip}
\hline
\noalign{\smallskip}
     Self-Monitoring + \texttt{ResNet-152}  & 13.34 & 4.02 & 0.62 & 0.62 \\
     Self-Monitoring + \texttt{pre-training} + \texttt{ResNet-152} & 12.3 & 3.03 & 0.63 & 0.7 \\
     Self-Monitoring + \texttt{blind}  & 15.64 & 7.1 & 0.23 & 0.32 \\
     Regretful-Agent + \texttt{ResNet-152} & 12.66 & 4.18 & 0.51 & 0.59 \\
     Regretful-Agent + \texttt{pre-training} + \texttt{ResNet-152} & 12.49 & 3.07 & 0.63 & 0.71 \\
     Regretful-Agent + \texttt{blind}  & 19.05 & 7.6 & 0.14 & 0.27 \\
\hline
\end{tabular}
\end{center}
\end{table}
\setlength{\tabcolsep}{1.4pt}

\setlength{\tabcolsep}{4pt}
\begin{table}
\begin{center}
\caption{Visual ablation study on Self-Monitoring \cite{Self-Monitoring} and Regretful-Agent \cite{regretful}, unseen environment with Single Run (not Beam Search).}
\label{table:blind-study-unseen}
\begin{tabular}{lllll}
\hline\noalign{\smallskip}
Model & PL $\downarrow$ & NE $\downarrow$ & SPL $\uparrow$ & SR $\uparrow$ \\
\noalign{\smallskip}
\hline
\noalign{\smallskip}
     Self-Monitoring + \texttt{ResNet-152}  & 15.88 & 6.47 & 0.27 & 0.39 \\
     Self-Monitoring + \texttt{pre-training} + \texttt{ResNet-152} & 16.27 & 5.99 & 0.30 & 0.42 \\
     Self-Monitoring + \texttt{blind}  & 15.86 & 6.6 & 0.24 & 0.35 \\
     Regretful-Agent + \texttt{ResNet-152} & 16.09 & 5.99 & 0.30 & 0.43 \\
     Regretful-Agent + \texttt{pre-training} + \texttt{ResNet-152} & 15.75 & 5.62 & 0.35 & 0.47 \\
     Regretful-Agent + \texttt{blind}  & 18.8 & 6.62 & 0.19 & 0.36 \\
\hline
\end{tabular}
\end{center}
\end{table}
\setlength{\tabcolsep}{1.4pt}

\setlength{\tabcolsep}{4pt}
\begin{table}
\begin{center}
\caption{Language ablation study on Self-Monitoring \cite{Self-Monitoring} and Regretful-Agent \cite{regretful}, unseen environment with Single Run (not Beam Search). w/ means without}
\label{table:text-study-unseen}
\begin{tabular}{lll}
\hline\noalign{\smallskip}
Model & PL $\downarrow$ & SR $\uparrow$ \\
\noalign{\smallskip}
\hline
\noalign{\smallskip}
    Training with real data\\
    
     Regretful-Agent \textit{baseline} & 16.1 & \textbf{0.43}  \\
     Regretful-Agent w/ \texttt{nouns} & 15.5 & 0.35 \\
     Regretful-Agent w/ \texttt{adjectives} & 14.8 & 0.42  \\
     Regretful-Agent w/ \texttt{nouns+adjectives}  & 14.9 & 0.37  \\
     Regretful-Agent w/ \texttt{all textual features} & 18.0 & 0.25  \\
    \noalign{\smallskip}
    \hline
    \noalign{\smallskip}
    Training with real data + augmented data\\
     
     Regretful-Agent \textit{baseline} & 15.8 & 0.47  \\
     Regretful-Agent w/ \texttt{nouns} & 14.8 & 0.36 \\
     Regretful-Agent w/ \texttt{adjectives} & 15.5 & \textbf{0.48}  \\
     Regretful-Agent w/ \texttt{nouns+adjectives}  & 13.9 & 0.39  \\
     Regretful-Agent w/ \texttt{all textual features} & 18.0 & 0.25  \\
\hline
\end{tabular}
\end{center}
\end{table}
\setlength{\tabcolsep}{1.4pt}

\subsection{Ablation studies}

When experimenting in the visual area by removing all the visual features we notice a clear difference between seen and unseen environments. In seen environments the difference in success rate is very large (Table \ref{table:blind-study-seen}). While the Self-Monitoring and Regretful-Agent models achieve about 60\% of success rate (SR) without pre-training, removing the agent's sight (+ \texttt{blind}) greatly reduces its performance (-30\%).

In unknown environments the difference is much smaller. We observe that both models without pre-training don't improve by more than 7\% SR over the blind model. This demonstrates good memorization but lack of generalization, being visual information almost useless on previously unknown scenes.

When experimenting in the linguistic area, we noticed that when we extract the whole language, it only reaches a 25\% SR. This means that 1 out of 4 random walks actually reaches the goal, noting the biases of the R2R dataset, where agents can navigate correctly to the goal point without any instruction.

If we extract the nouns or nouns+adjectives from the instruction, then the model reduces the SR moderately. This explains that many of the instructions are based on prompts such as ``turn right'' or ``walk straight to the bottom'', without necessarily reference the environment. 

Because of the high-level instructions, removing adjectives increases the SR, indicating that nouns descriptions are actually interfering with the model performance. 

We propose to create and train with semantically richer instructions, in order to include more detailed description of the visual environment and then force the agent to use all the available information.

\section{Semantically richer instructions proposal} \label{section:object-based}

To navigate using vision, we must first learn to follow semantically meaningful instructions that foster the use of visual information. We create simple instructions that are scene-object based, referencing them and their context, enriching over generic non visual instructions like ``go straight".

\begin{figure*}[t]
    \centering
    \includegraphics[width=0.6\linewidth]{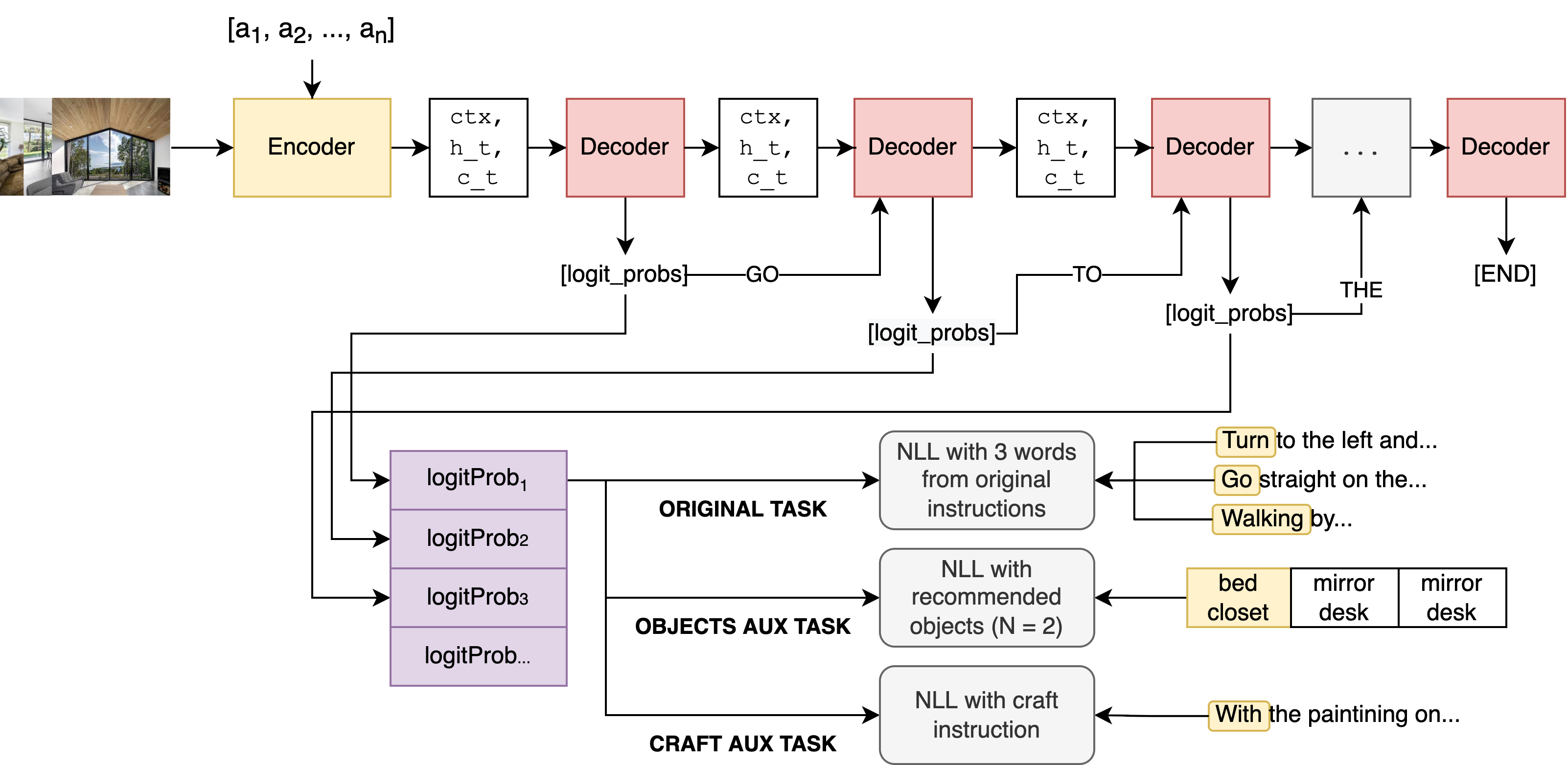}
    \caption{Speaker with proposed auxiliary tasks.}
    \label{fig:speaker_with_objects}
\end{figure*}

We construct our model over Speaker-Follower, but using only the Speaker module. On Figure \ref{fig:speaker_with_objects} is our complete model. The original Speaker module takes, for each path, the sequence of panoramic views and also the actions sequence (\texttt{RIGHT}, \texttt{<END>}, \texttt{FORWARD}, etc.), and pass them across an encoder module. This encoder gives us an encoded context \texttt{ctx}, used for generating each word of the new instruction through an LSTM, which uses also the previous cell and hidden states, as shown in the figure. 

On R2R, each path has three instructions. For each, the Speaker module builds the loss as the Negative Log Loss (NLL) between the corresponding word of the instruction and the generated word, as shown on the figure. 

Generated instructions on the Speaker module are now being used for almost all state-of-the-art models of VLN task on a pre-training phase. However, it has been shown that they do not follow human syntax, they have orientation problems and do not include relevant information, being incorrect in most cases \cite{zhao2021evaluation}. 

Using Matterport3DMeta, we propose to add relevant objects to generated instructions applying two loss auxiliary tasks: objects and crafted instructions, aiming the vision to be mandatory for the agent to navigate. These two are also included in Figure \ref{fig:speaker_with_objects}.

\subsection{Objects and crafted instructions}

Object metadata is available on Matterport3DMeta, but it is raw and difficult to use. That's why we created \texttt{360-visualization}, a script for fetching and visualizing objects and navigable viewpoints on each node, for each heading and elevation.

These objects are the main component of the objects auxiliary task, but we also use them for the generation of crafted instructions. 

\begin{figure*}[h]
    \centering
    \includegraphics[width=\linewidth]{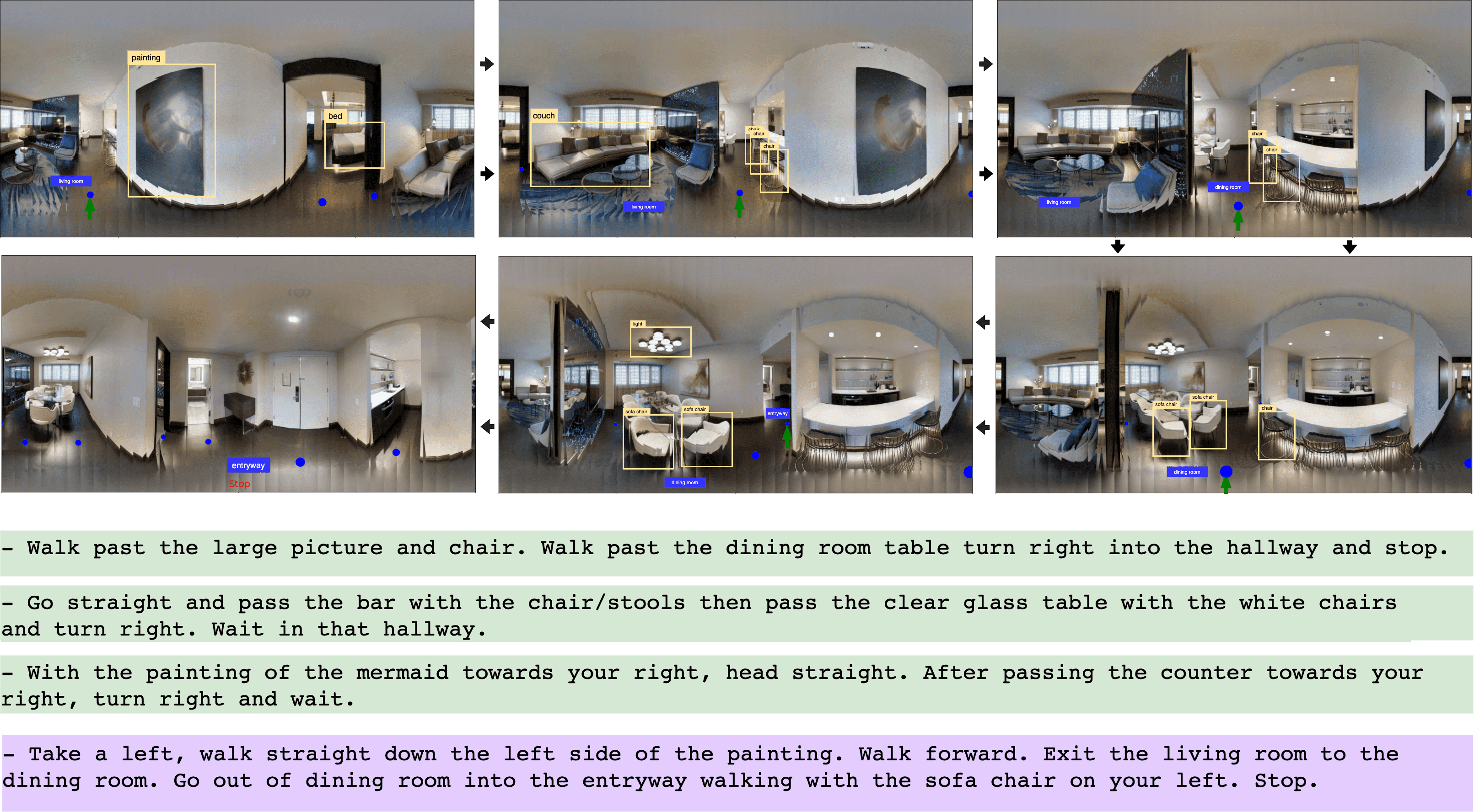}
    \caption{Crafted instruction example. Panoramic views sequence on top and \textcolor{applegreen}{human instructions} + \textcolor{brightlavender}{crafted instruction} on bottom. Images are sequenced through the arrows. Presented objects and scenes names are sampled from the data.}
    \label{fig:craft-example}
\end{figure*}

For a specific path, we generate an atomic instruction for each node on the sequence. Having the current 360\,° visual image and the next node we can select the best object to reference, following a set of rules. For instance, in Figure \ref{fig:craft-example} we start with a big painting at the right of the next node, generating the first atomic instruction: ``Turn left, walk straight down the left of the painting.". Then, we concatenate all this atomic instructions, generating a new crafted instruction for the selected path.

\subsection{Objects auxiliary task}

For each node of a path sequence, we fetch all objects with \texttt{360-visualization} and filter them by distance, area, uniqueness and usability (excluding many objects, like ``floor" that has large area). 

We assign the best \texttt{N} objects to each word of the instructions of that path, matching word index with the closer node index.

For instance, in Figure \ref{fig:speaker_with_objects} we recommend the model to use ``bed" and ``closet" (\texttt{N = 2}) for the first word.

\begin{figure*}[h]
    \centering
    \includegraphics[width=1
    \linewidth]{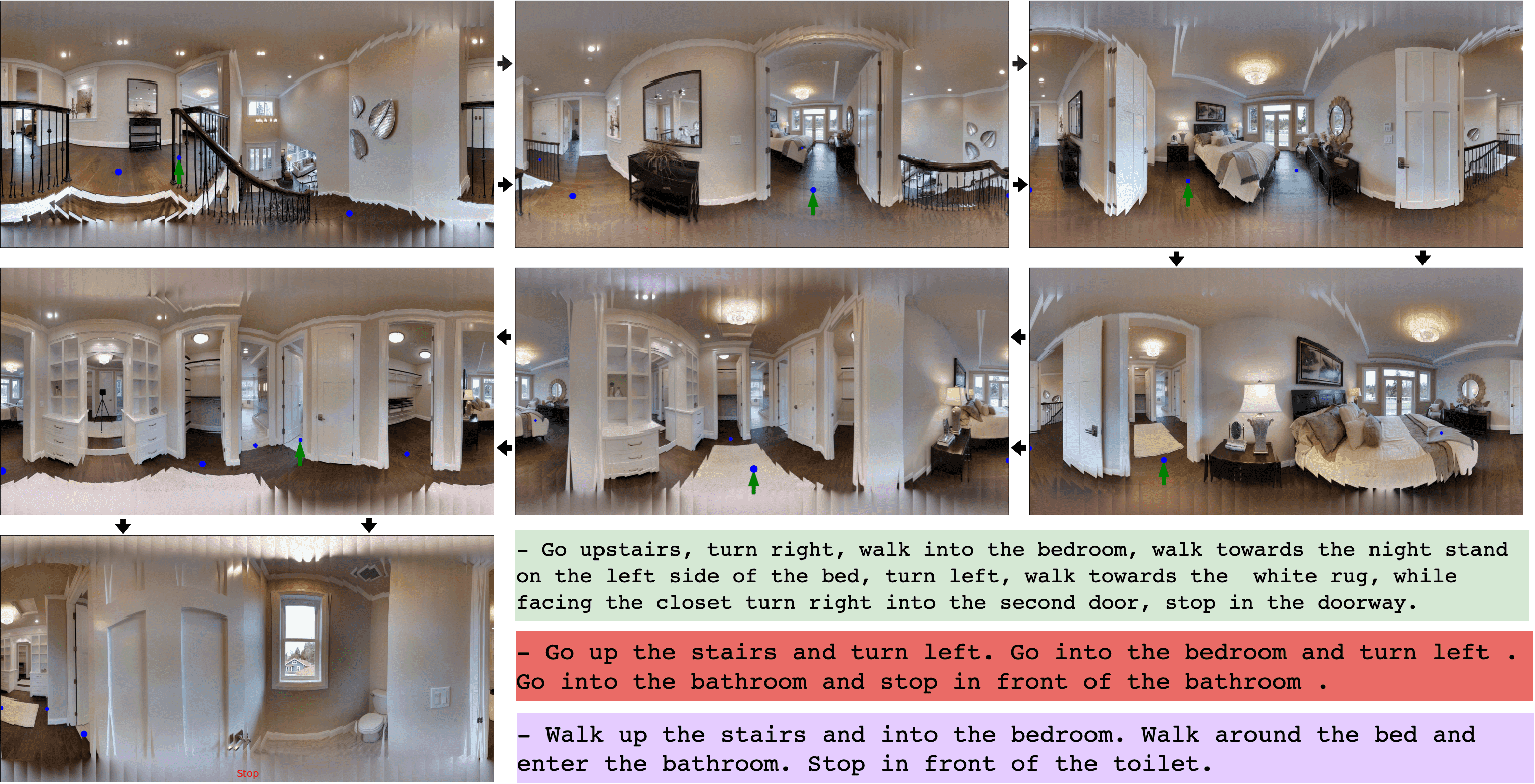}
    \caption{Objects auxiliary task instruction example. Panoramic views sequence above and \textcolor{applegreen}{human instruction} + \textcolor{brickred}{base speaker instruction} + \textcolor{brightlavender}{new generated instruction} below. Images are sequenced through the arrows.}
    \label{fig:objs-example}
\end{figure*}

The objects auxiliary task consists on adding Negative Log Loss between the generated word and the \texttt{N} recommended objects to the final loss. The sum of these losses are weighted by $\lambda$, a modifiable parameter. 

The final loss on training the Speaker results as follows
$$ wordLoss = \sum_{i=0}^3 NLL \left( log(logit), w_{original_i} \right) + \lambda \sum_{i=0}^N NLL \left( log(logit), w_{object_i} \right) $$

A resulting generated instruction with Speaker + objects auxiliary task is shown on Figure \ref{fig:objs-example}. 

\subsection{Crafted instructions auxiliary task}
Having our own crafted instructions, we use them directly on training and also adding a Negative Log Loss between the generated instruction and the crafted instruction, word by word. 

The sum of these losses are weighted by $\beta$, another modifiable parameter. The final loss for each generated word is $$ wordLoss = \sum_{i=0}^3 NLL \left( log(logit), w_{original_i} \right) + \beta \cdot NLL \left( log(logit), w_{crafted} \right) $$

\subsection{Results and discussion}

After training the Speaker module with auxiliary tasks, we generate instructions based on the same sampled paths as the original augmented dataset with different values for $\lambda$ and $\beta$, hyperparameters that weigh auxiliary tasks loss. We execute separated auxiliary tasks, because it is redundant information. Crafted instructions used in the crafted instructions auxiliary task use the same objects as the ones we pass directly on the objects auxiliary task.

We then train the Regretful Agent navigation model with the augmented data or crafted instructions directly (Phase I), and finetune it with the original training data (Phase II). Results for seen and unseen environments are on Tables \ref{table:results-aux-task-seen} and \ref{table:results-aux-task-unseen}.

Best results are achieved pre-training with instructions generated using objects auxiliary task with $\lambda = 0.5$ or crafted instructions auxiliary task with $\beta = 0.3$, increasing up to 51\% the success rate on unseen environment. Using our new augmented instructions the success rate increment is doubled compared using original Speaker module synthetic instructions, this is, +8\% versus the baseline model.

We extend the experiments to the HAMT model \cite{chen2021hamt}, pre-training with the best settings we got from above. Results are on Table \ref{table:hamt-results-aux-task-unseen}. Although we get a lower increase, adding visual information to the instructions is still useful across all models. As future work, we can test with different configurations of the auxiliary tasks to see which dataset generates higher benefits. We make objects and crafted instructions available in \texttt{360-visualization}\footnote{\url{https://github.com/cacosandon/360-visualization}} so they can be added to any language model, as we propose with the Speaker module.

The new module with auxiliary tasks have several advantages in the quality of the generated instructions. One example is shown on Figure \ref{fig:objs-example}. 

First, it corrects the Speaker module in orientation, since we help the model by indicating which objects to reference in the instruction, relating them with the next node position. For instance, in the figure's sequence the agent must turn right, while the Speaker generates an instruction that wrongly says the opposite.

Second, compared to human instructions we realize that a complex instruction is not necessary, since agent is a beginner on navigation. Humans describe the path in a high level, which makes it even more complicated to extract visual information. Our module generates low-level instructions, describing the best way to orientate another DL model.

At last, generated instructions reference objects that exist in the environment but not in the original instructions, as it does with the word ``toilet" at the end of the instruction showed in Figure \ref{fig:objs-example}. The model learned to use objects even though it has never seen them before, nor has any type of information (such as the instruction showed on the figure, which is from the validation set, and therefore, never seen before).

We also demonstrate that the increase of success rate using original data augmentation does not depend on the quality of the Speaker's instructions, but rather the quantity is the main contributor. This means, quantity compensates quality for improving performance (178,300 augmented instructions over 4675 train instructions). 

As we mentioned before, using the same 178,300 sampled paths, we also create totally new crafted instructions (Figure \ref{fig:craft-example}) without the need of human instructions in order to use them as pre-training. On Tables \ref{table:results-aux-task-seen} and \ref{table:results-aux-task-unseen} it is shown as ``\texttt{PWIF Crafted Directly}". We almost reach the same success rate as the full training (Phases I + II) with human instructions and we exceed base training by 4\%.

We then show that the Speaker module as a instruction generator does not contribute more than our crafted instructions generated based on rules, unless we add the proposed auxiliary tasks, where the performance increase is remarkable.

\setlength{\tabcolsep}{4pt}
\begin{table}
\begin{center}
\caption{Regretful-Agent \cite{regretful} pre-training with instructions generated from different models, evaluated on \textbf{seen} environments with Single Run (not Beam Search) and original human instructions. \texttt{PWIF} means \textit{Pre-training with instructions from} and \texttt{AT} means \textit{Auxiliary Task}.}
\label{table:results-aux-task-seen}
\begin{tabular}{lllll}
\hline\noalign{\smallskip}
Regretful-Agent + & PL $\downarrow$ & NE $\downarrow$ & SPL $\uparrow$ & SR $\uparrow$ \\
\noalign{\smallskip}
\hline
\noalign{\smallskip}
     \texttt{Without pre-training} & 12.66 & 4.18 & 0.51 & 0.59 \\
     \texttt{PWIF Speaker base} & 12.49 & 3.07 & 0.63 & 0.71 \\
     \noalign{\smallskip}
     \hline
     \noalign{\smallskip}
     \texttt{PWIF Speaker + Objects AT $\lambda = 0.3, N=2$} & 12.97 & 3.10 & 0.62 & 0.71 \\
     \texttt{PWIF Speaker + Objects AT $\lambda = 0.5, N=1$} & 11.65 & 3.38 & 0.61 & 0.67 \\
     \texttt{PWIF Speaker + Objects AT $\lambda = 0.5, N=2$} & 12.09 & 2.93 & 0.65 & \textbf{0.72} \\
     \texttt{PWIF Speaker + Objects AT $\lambda = 0.5, N=3$} & 12.80 & 3.48 & 0.58 & 0.67 \\
     \texttt{PWIF Speaker + Objects AT $\lambda = 0.6, N=2$} & 12.07 & 3.09 & 0.63 & 0.70 \\
     \texttt{PWIF Speaker + Crafted AT $\beta = 0.1$} & 12.41 & 3.16 & 0.62 & 0.70 \\
     \texttt{PWIF Speaker + Crafted AT $\beta = 0.2$} & 11.83 & 3.29 & 0.62 & 0.68 \\
     \texttt{PWIF Speaker + Crafted AT $\beta = 0.3$} & 12.24 & 2.86 & 0.63 & \textbf{0.72} \\
     \texttt{PWIF Speaker + Crafted AT $\beta = 0.4$} & 12.16 & 3.08 & 0.62 & 0.70 \\
     \texttt{PWIF Crafted directly} & 12.50 & 3.32 & 0.59 & 0.67 \\
\hline
\end{tabular}
\end{center}
\end{table}
\setlength{\tabcolsep}{1.4pt}

\setlength{\tabcolsep}{4pt}
\begin{table}
\begin{center}
\caption{Regretful-Agent \cite{regretful} pre-training with instructions generated from different models, evaluated on \textbf{unseen} environments with Single Run (not Beam Search) and original human instructions. \texttt{PWIF} means \textit{Pre-training with instructions from} and \texttt{AT} means \textit{Auxiliary Task}.}
\label{table:results-aux-task-unseen}
\begin{tabular}{lllll}
\hline\noalign{\smallskip}
Regretful-Agent + & PL $\downarrow$ & NE $\downarrow$ & SPL $\uparrow$ & SR $\uparrow$ \\
\noalign{\smallskip}
\hline
\noalign{\smallskip}
     \texttt{Without pre-training} & 16.09 & 5.99 & 0.30 & 0.43 \\
     \texttt{PWIF Speaker base} & 15.75 & 5.62 & 0.35 & 0.47 \\
     \noalign{\smallskip}
     \hline
     \noalign{\smallskip}
     \texttt{PWIF Speaker + Objects AT $\lambda = 0.3, N=2$} & 15.27 & 5.39 & 0.36 & 0.49 \\
     \texttt{PWIF Speaker + Objects AT $\lambda = 0.5, N=1$} & 14.66 & 5.80 & 0.35 & 0.46 \\
     \texttt{PWIF Speaker + Objects AT $\lambda = 0.5, N=2$} & 14.61 & 5.29 & 0.39 & \textbf{0.51} \\
     \texttt{PWIF Speaker + Objects AT $\lambda = 0.5, N=3$} & 15.24 & 5.77 & 0.34 & 0.47 \\
     \texttt{PWIF Speaker + Objects AT $\lambda = 0.6, N=2$} & 15.82 & 5.46 & 0.34 & 0.48 \\
     \texttt{PWIF Speaker + Crafted AT $\beta = 0.1$} & 14.90 & 5.75 & 0.35 & 0.47 \\
     \texttt{PWIF Speaker + Crafted AT $\beta = 0.2$} & 14.37 & 5.58 & 0.38 & 0.48 \\
     \texttt{PWIF Speaker + Crafted AT $\beta = 0.3$} & 15.42 & 5.52 & 0.37 & 0.50 \\
     \texttt{PWIF Speaker + Crafted AT $\beta = 0.4$} & 15.44 & 5.43 & 0.36 & 0.47 \\
     \texttt{PWIF Crafted directly} & 15.97 & 6.03 & 0.33 & 0.46 \\
\hline
\end{tabular}
\end{center}
\end{table}
\setlength{\tabcolsep}{1.4pt}

\setlength{\tabcolsep}{4pt}
\begin{table}
\begin{center}
\caption{HAMT \cite{chen2021hamt} pre-training with instructions generated from best Speaker configurations (ranked by success rate after pre-training the Regretful Agent), evaluated on \textbf{unseen} environments with Single Run (not Beam Search) and original human instructions. \texttt{PWIF} means \textit{Pre-training with instructions from} and \texttt{AT} means \textit{Auxiliary Task}.}
\label{table:hamt-results-aux-task-unseen}
\begin{tabular}{lllll}
\hline\noalign{\smallskip}
HAMT + & SPL $\uparrow$ & SR $\uparrow$ \\
\noalign{\smallskip}
\hline
\noalign{\smallskip}
     \texttt{Without pre-training} & 54.4 & 48.7 \\
     \texttt{PWIF Speaker base} & 56.3 & 52.3 \\
     \noalign{\smallskip}
     \hline
     \noalign{\smallskip}
     \texttt{PWIF Speaker + Objects AT $\lambda = 0.5, N=2$} & \textbf{57.4} & 52.4 \\
     \texttt{PWIF Speaker + Crafted AT $\beta = 0.3$} & 57.3 & \textbf{52.6} \\
\hline
\end{tabular}
\end{center}
\end{table}
\setlength{\tabcolsep}{1.4pt}

\section{Conclusions} \label{conclusions}

Different methodologies have been developed to improve scene understanding, in order to achieve a better performance in navigation with human interaction. They focus mainly on model architecture, leaving aside the base of the task: the dataset. As we present, navigation agents do not use visual information available on environments for making a decision. Removing visual features generates a slight success rate drop of only 7\% on unseen environments, evidencing that the R2R dataset has instructions that do not reference the context in which agent is situated. This last factor allows agents to execute actions in almost a random manner, reaching the goal anyway. 

In addition, these same instructions are too complex and high level, confusing agents that start as beginners on navigation. To bridge the visual semantic gap presented on the datasets, we create new semantically richer instructions.

For this purpose, we use scene objects and crafted instructions to feed a set of auxiliary tasks. The resulting model generates new instructions that help to correct the errors existing in the original instructions, while increase the success rate by 8\% in unseen environments when we use them as pre-training, which doubles the increase of the original Speaker. We then demonstrate that the creation of semantically richer instructions that include explicit visual information allows the agent to better learn to navigate.

In order to follow this same line to improve robot navigation, we propose different branches for further research:
\begin{itemize}
    \item \textbf{Own object detection}: We construct our auxiliary task based on available metadata of different environments (Matterport3DMeta). If we want to expand to new environments where this metadata is not available, we must detect objects on our own. Indoor object detection is an unresolved task, which can be improved directly using the same scene objects that we retrieve from raw data. 
    
    \item \textbf{3-phase Curriculum Learning}: We pre-train our model with our semantically richer instructions, and then finetune with the original instructions, which are complex and high level. Starting with an easier task will allow the agent to use environment information progressively. Standing in a random node, we have the 360\,° image, different possible navigation nodes and an atomic instruction. The agent has to decide which node to move to. The agent will learn simpler and shorter instructions that refer to the environment, the basics for starting to execute this tasks on sequence.
\end{itemize}

\section*{Acknowledgments} This work was partially funded by FONDECYT grant 1221425, the National Center for Artificial Intelligence CENIA FB210017, Basal ANID and by ANID through \textit{Beca de Magister Nacional} N° 22210030. 

\clearpage
% ---- Bibliography ----
%
% BibTeX users should specify bibliography style 'splncs04'.
% References will then be sorted and formatted in the correct style.
%
\bibliographystyle{splncs04}
\bibliography{egbib}
\end{document}